\documentclass[10pt,twocolumn,letterpaper]{article}
\usepackage{iccv}
\usepackage{times}
\usepackage{epsfig}
\usepackage{graphicx}
\usepackage{amsmath}
\usepackage{amssymb}
\usepackage{subfigure}
\usepackage{float}

\usepackage[colorlinks, bookmarks=ture]{hyperref}

\iccvfinalcopy 

\def\iccvPaperID{11883} 

\ificcvfinal\fi

\begin{document}

\title{DEYOv2: Rank Feature with Greedy Matching for End-to-End Object Detection}

\author{Haodong Ouyang\\
Southwest Minzu University\\
Chengdu, China\\
{\tt\small ouyanghaodong@stu.swun.edu.cn}
}

\maketitle
\ificcvfinal\thispagestyle{plain}
\setlength{\textfloatsep}{0pt}

\begin{abstract}
This paper presents a novel object detector called DEYOv2, an improved version of the first-generation DEYO $($DETR with YOLO$)$ model. DEYOv2, similar to its predecessor, DEYOv2 employs a progressive reasoning approach to accelerate model training and enhance performance. The study delves into the limitations of one-to-one matching in optimization and proposes solutions to effectively address the issue, such as Rank Feature and Greedy Matching. This approach enables the third stage of DEYOv2 to maximize information acquisition from the first and second stages without needing NMS, achieving end-to-end optimization. By combining dense queries, sparse queries, one-to-many matching, and one-to-one matching, DEYOv2 leverages the advantages of each method. It outperforms all existing query-based end-to-end detectors under the same settings. When using ResNet-50 as the backbone and multi-scale features on the COCO dataset, DEYOv2 achieves 51.1 AP and 51.8 AP in 12 and 24 epochs, respectively. Compared to the end-to-end model DINO, DEYOv2 provides significant performance gains of 2.1 AP and 1.4 AP in the two epoch settings. To the best of our knowledge, DEYOv2 is the first fully end-to-end object detector that combines the respective strengths of classical detectors and query-based detectors.
\end{abstract}

\vspace{-7mm}
\section{Introduction}
Identifying regions of interest in an image and labeling them with bounding boxes and class labels is a crucial task in computer vision known as object detection. This task has numerous applications and has been improved significantly with the development of deep learning. Over the past few decades, several exceptional one-stage and two-stage object detection models have been developed. The R-CNN \cite{54,9, 16} family is the most well-known two-stage object detector, which includes Fast R-CNN\cite{16} and Faster R-CNN\cite{9}. Meanwhile, YOLO\cite{4,5,6}, SSD\cite{20}, and RetinaNet\cite{21} are the most popular one-stage object detec-
\begin{figure}[H]
\vspace{3mm}
\begin{center}
\includegraphics[width=0.75\linewidth]{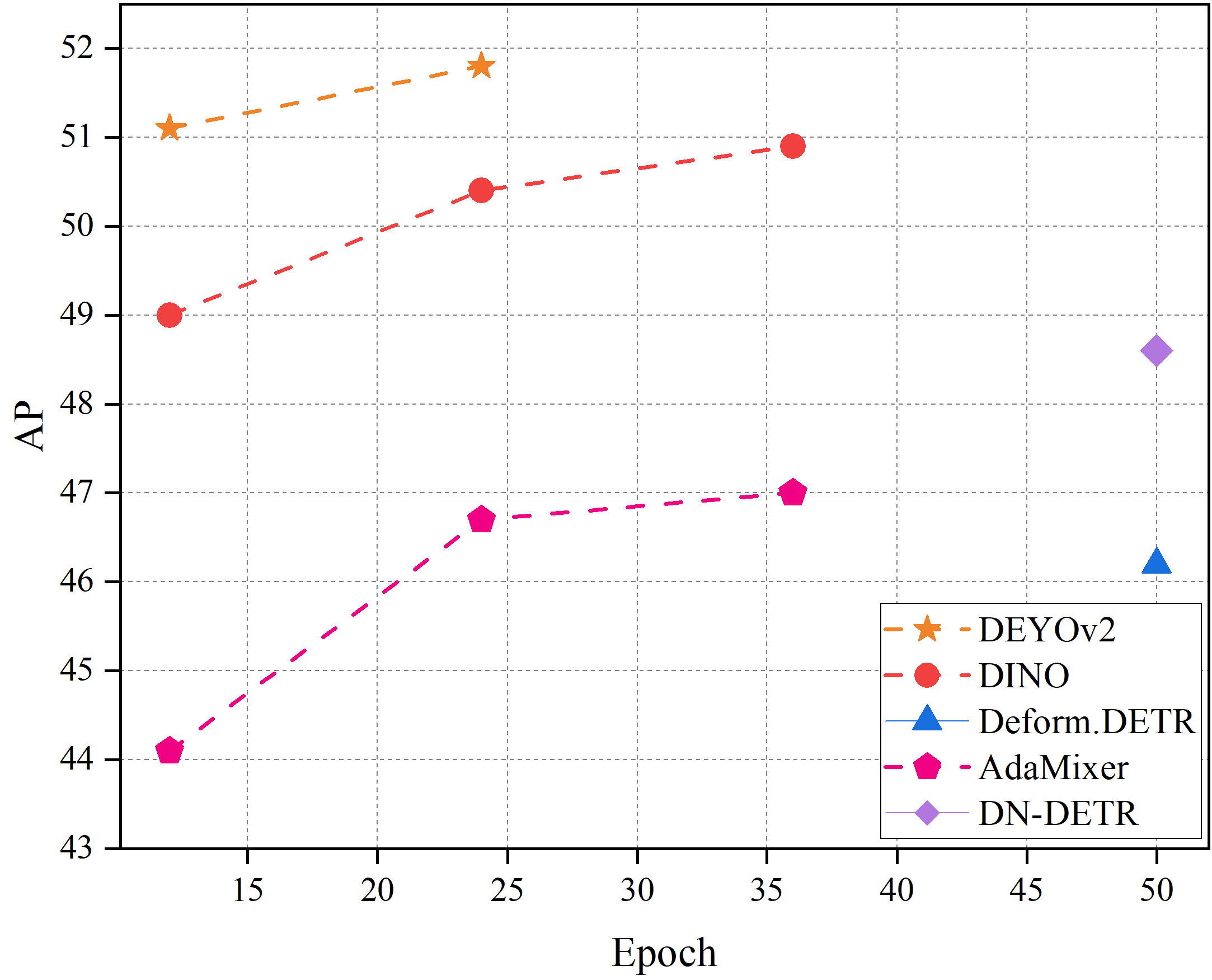}
   \caption{We have conducted a comparative analysis of various end-to-end object detectors, wherein we have evaluated their performance on the COCO 2017 validation dataset by using a ResNet50 backbone.}
\end{center}
\label{fig:1}
\vspace{-6mm}
\end{figure}
\noindent tor models. Classical detectors have one thing in common: their heavy reliance on hand-crafted components, such as non-maximum suppression (NMS). Because these detection algorithms usually output multiple candidate bounding boxes, each corresponding to an area where an object may exist. However, there are often overlaps or redundancy between these candidate boxes, which need to be screened and optimized.

Although NMS is a useful algorithm for object detection, it does have some limitations. One issue is that it may inadvertently remove bounding boxes overlapping significantly with the highest-scoring bounding box, especially in areas with dense or similarly-sized objects. Additionally, the effectiveness of NMS can be impacted by the chosen IoU threshold, potentially resulting in significant changes to detection results. NMS may not affect detector recall in sparser scenarios, but it can become a performance bottleneck for classical detectors in crowded scenes.

\begin{figure*}[t]
\begin{center}
\includegraphics[width=0.8\linewidth]{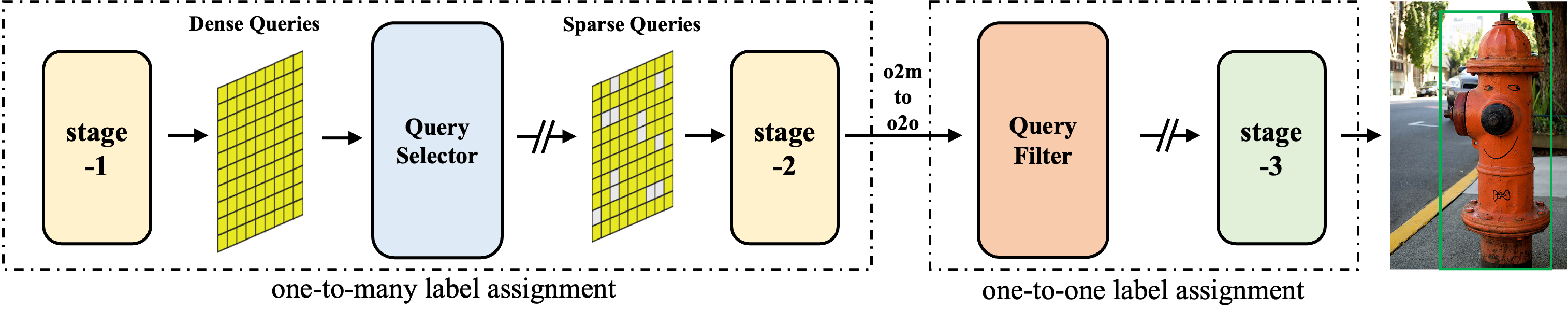}
   \caption{{\bf The pipeline of DEYOv2}. Stage1 provides high-quality dense queries for stage2 and stage3. stage2 uses the greedy matching and rank feature to establish the score gap between each object so that the query filter can filter out the redundant bounding box and provide a guarantee for the optimization of stage3. -//- means stop gradient back-propagation.
}
\end{center}
\label{fig:2}
\vspace{-5mm}
\end{figure*}

The Detection Transformer (DETR)\cite{1} presents an innovative transformer-based object detector that utilizes a transformer-based encoder-decoder framework. Instead of relying on manual components of NMS, DETR\cite{1} uses the Hungarian loss to predict one-to-one object sets, resulting in end-to-end optimization. NMS does not perform well in crowded scenes because it only clusters objects based on simple information like IoU and categories, possibly grouping different objects of similar size into one group. In contrast, DETR\cite{1} leverages the interaction between queries to utilize more complex information and distinguish the relationship between them. This approach makes DETR's strategy of predicting one-to-one object sets more reasonable compared to NMS, resulting in significantly better performance in crowded scenes than classical detectors that use NMS.

 Although DETR\cite{1} has attracted great interest from the research community, it also has many problems. Firstly, it should be noted that DETR\cite{1} has a slow convergence rate, requiring 500 training epochs to achieve acceptable performance. However, the DEYO\cite{52} algorithm presents a new perspective on improving DETR\cite{1}, taking inspiration from the idea of Step-by-Step. DEYO\cite{52} uses low-cost and high-quality YOLO\cite{4,5,6} predictions as the input of the second-stage DETR-like model to reduce the difficulty level of the DETR-like model for predicting one-to-one object sets. DEYO\cite{52} combines classical and query-based detectors with their respective advantages, thereby improving overall performance. At the same time, DEYO\cite{52} also discovered the limitations of one-to-one label assignment. Since DETR uses one-to-one matching, the strategy of establishing a score gap is adopted to suppress redundant bounding boxes. This means that the decoder needs to model the relationship between queries and distinguish between optimal and redundant bounding boxes. The experiment of DEYO\cite{52} shows that the decoder is very weak in dealing with a large number of almost similar redundant bounding boxes. As the congestion of proposal queries increases, the performance gradually decreases, and even training crashes when NMS is not used. As shown in Figure 3, as the IoU threshold rises, meaning that queries with increased congestion are fed into the decoder, the performance also drops, and the AP plummets to 0 when NMS is not used. And we find that this performance drop is not mitigated by using NMS with a low IoU threshold in the post-process. This shows that it is not only the redundant bounding box that affects the final performance, but the redundant bounding box also seriously hinders the training and optimization of the detector during training. The above phenomenon shows that the filtering ability of query-based detectors is limited, its suppression strategy for bounding boxes is not good, and there are strict requirements on query initialization, which limit the design of query-based detectors to some extent. DEYO\cite{52} simply solves this problem by using NMS to filter out redundant bounding boxes, but as shown in Figure 9, the effective information of stage1 loses more due to the decrease of NMS IoU threshold, which not only destroys the end-to-end advantage of DETR but also limits the effective information transmission of stage1. Just like classical detectors, this problem is undoubtedly exacerbated in crowded scenes.

\begin{figure}[t]
\begin{center}
\includegraphics[width=0.7\linewidth]{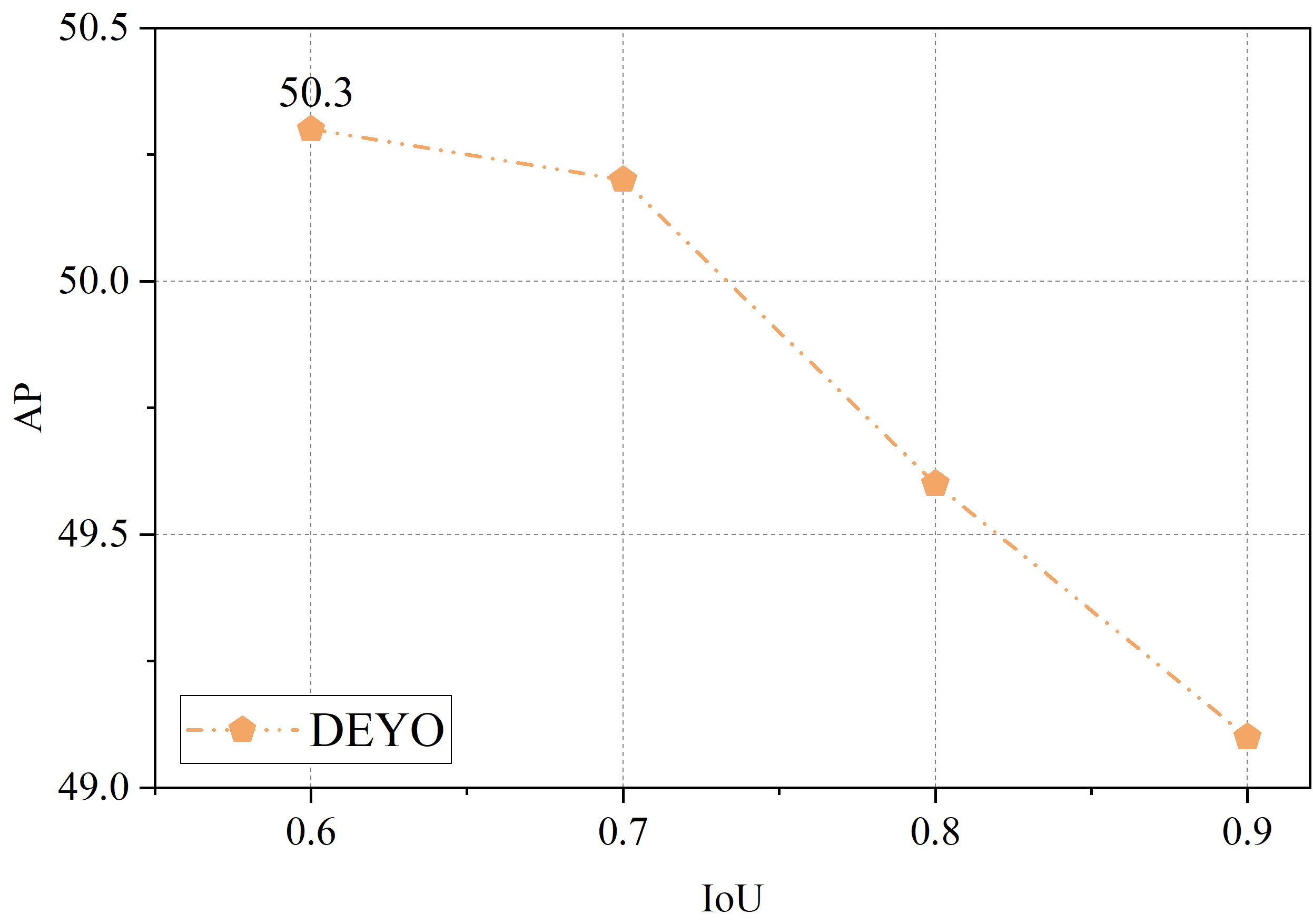}
   \caption{One-to-one matching is very weak in dealing with redundant boxes, and the final performance of the detector decreases with the increase of the crowding degree of proposal queries.
}
\end{center}
\label{fig:3}
\end{figure}

So is there a way to reduce information loss, achieve more elegance, and avoid optimization caused by redundant bounding boxes? DEYOv2 solves the above problems by introducing rank feature and greedy matching. The rank feature refers to the feature of the rank after the bounding box is sorted by confidence and is encoded by embedding. Adding it to the query of the decoder can make it easy for the detector to learn the strategy of non-maximum value suppression so that the detector is faced with a crowded environment. The bounding box can still maintain a good filtering effect. Greedy matching can use no additional labels, use the ground truth as a benchmark, cluster and supervise each bounding box, and guide the model to select the bounding box with the largest rank from each class to retain. rank feature and greedy matching enable DEYOv2 to get rid of the dependence on NMS in the transition from one-to-many label assignment to one-to-one label assignment, and it solves the optimization problem encountered by the Transformer encoder when filtering redundant bounding boxes to achieve end-to-end optimization.

In this paper, we propose a new paradigm of the three-stage object detection network called DEYOv2, to our best knowledge, is the first fully end-to-end detector that combines the respective advantages of classical detectors and query-based object detectors. Thus our DEYOv2 becomes a new state-of-the-art for end-to-end object detector. As shown in Figure 1, under the setting of ResNet50 4scales, we have achieved the best performance of 51.1 AP and 51.8 AP on the COCO2017 val dataset\cite{12} at 12 epochs and 24 epochs, respectively.

\begin{figure}[t]
\begin{center}
\includegraphics[width=0.65\linewidth]{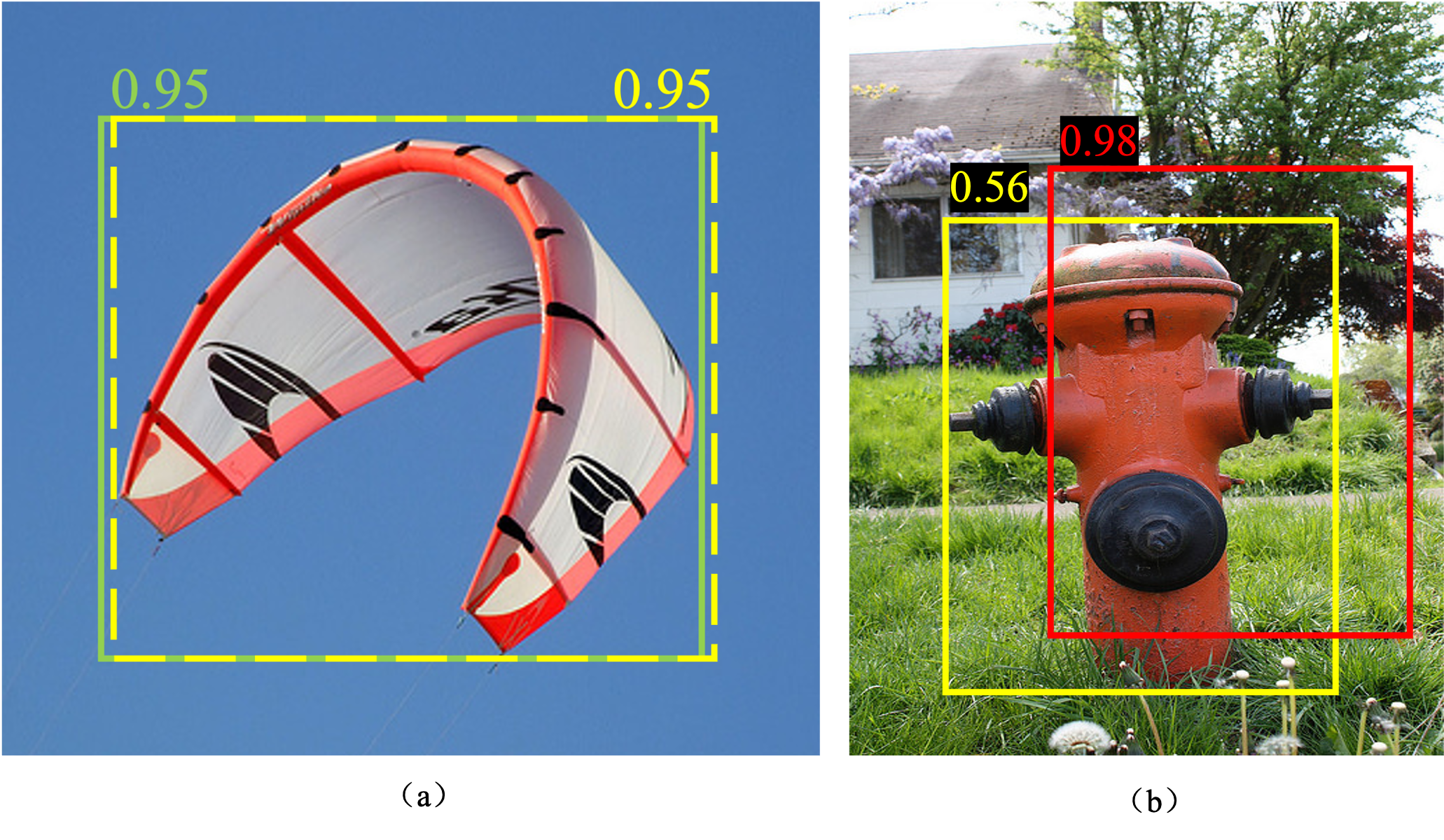}
   \caption{(a) shows the situation where one-to-many matching cannot make a choice between two nearly similar objects. (b) shows that high confidence does not always mean good positioning.
}
\end{center}
\label{fig:4}
\end{figure}

The main contributions of this paper are summarized as follows: (i) We propose the first fully end-to-end object detector that combines the respective advantages of classical detectors and query-based object detectors, which not only outperforms the current best end-to-end detector and does not need to rely on the manual component NMS to filter out the redundant bounding boxes of stage1, so the inference speed is not delayed and remains stable; (ii) We deeply analyzed the impact of redundant bounding boxes on one-to-one matching optimization, and proposed rank feature and greedy matching to use a more reasonable strategy to filter out a large number of dense redundant bounding boxes. (iii) We conducted several experiments verifying our ideas and exploring the contribution of each component within our model.

\section{Motivation}
Combined with the previous discussion, we believe that DETR's decoder is quite successful in establishing the relationship between each bounding box because compared with crowded bounding boxes, it should be more difficult to distinguish between sparse bounding boxes, and DEYO\cite{52} has achieved good performance on the low IoU threshold, which shows that DETR\cite{1} can handle the clustering of sparse bounding boxes well, so DETR\cite{1} should also be able to handle the clustering of crowded bounding boxes well, the difficulty of the latter is obviously lower than the former. This inspires us to re-examine the current filtering strategy, analyze why the crowded bounding box hinders the optimization of the detector, and propose an effective solution. We guess that for bounding boxes that are almost similar bounding boxes, although the detector can cluster them well, it is difficult to learn an effective strategy to select a bounding box from the cluster and keep it. In extreme cases, for n identical bounding boxes, using one-to-one label assignment, it is impossible for DETR\cite{1} to learn an effective filtering strategy from the original data. At the same time, when the bounding box is too similar, it may also lead to the instability of binary matching. Further hinder optimization. DEYO\cite{52} simply uses NMS to avoid optimization problems by establishing the IoU gap. We think that the establishment of the score gap should also achieve similar results.

\begin{figure}[t]
\begin{center}
\includegraphics[width=0.98\linewidth]{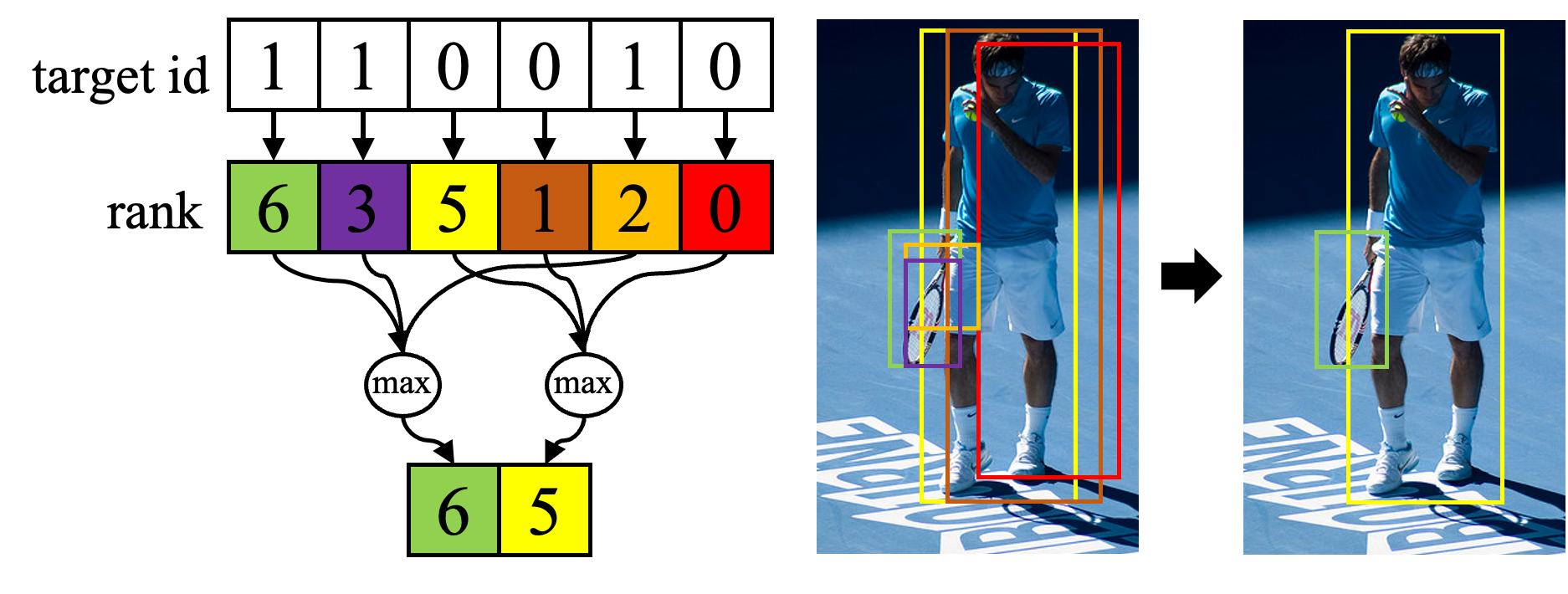}
   \caption{We assign the closest ground truth to each object and select the largest rank from each group to keep.
}
\end{center}
\label{fig:5}
\end{figure}
\section{Method}

\begin{figure*}[t]
\begin{center}
\includegraphics[width=0.7\linewidth]{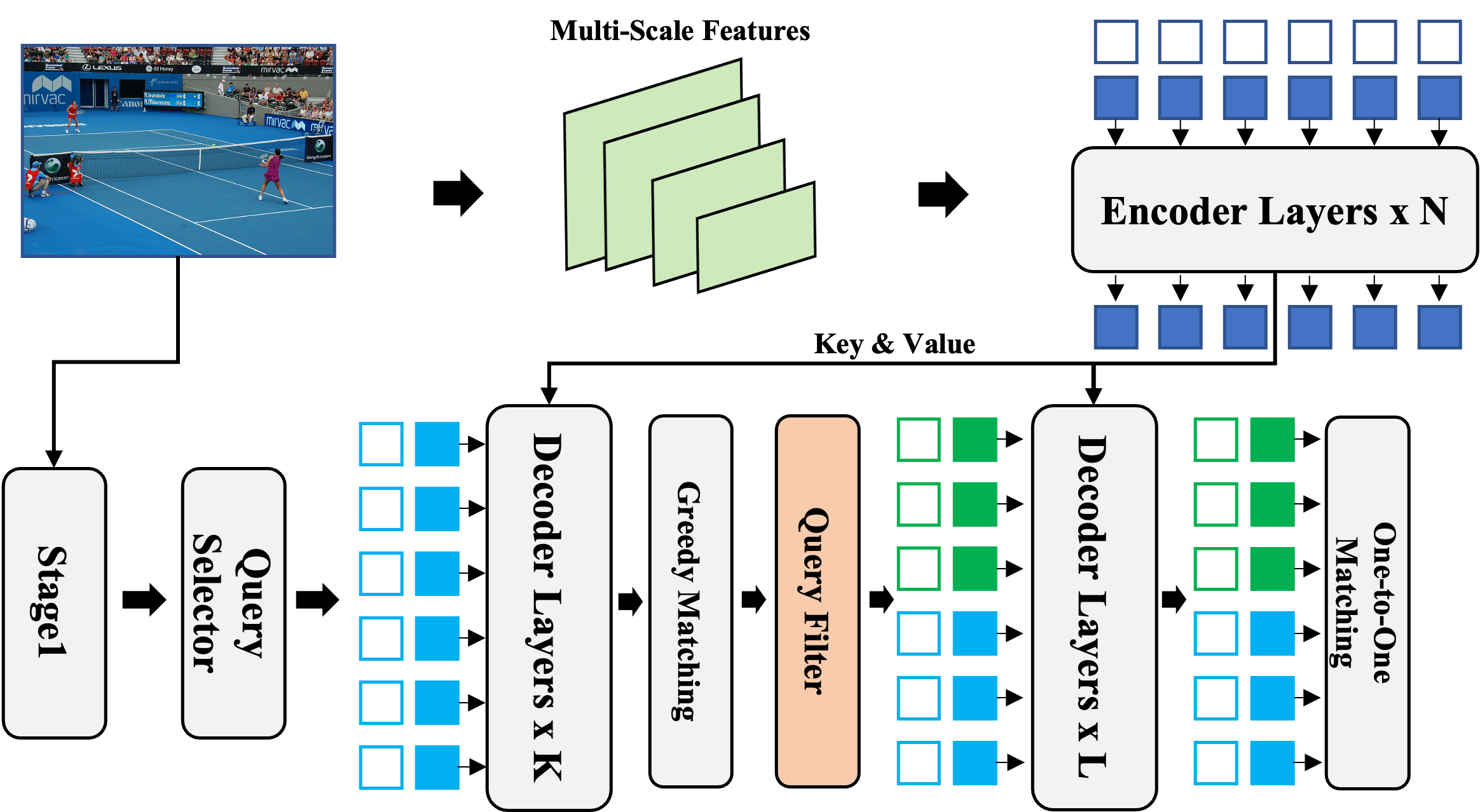}
   \caption{The framework of our proposed DEYOv2 model. Our improvements are mainly in stage2 and stage3. We simply use top-k to filter the information from stage1 to complete the transition from dense query to sparse query. We use the rank feature and greedy matching to create a score gap for proposal queries from stage1, which is convenient for query filter to filter out redundant bounding boxes, thus solving stage3 optimization problems.
}
\end{center}
\label{fig:6}
\vspace{-7mm}
\end{figure*}

\subsection{Rank Feature}
As shown in Figure 4, when encountering almost the same bounding box, it is difficult for the detector to learn an effective strategy to select a bounding box to keep, and this is exactly what NMS is good at. The non-maximum value strategy can make NMS selects a bounding box from a large number of nearly identical redundant bounding boxes without causing performance loss. We tried several strategies to let the model learn the strategy of non-maximum suppression directly from the original data, but all ended in failure. In order to reduce the training difficulty of the model, inspired by \cite{56}, we introduce the rank feature to solve this problem. It is found that adding the rank feature performs better than directly passing the confidence to the model. We think the reason for better performance is that the rank feature can make it easy for the model to learn the strategy of non-maximum value suppression. This is because the rank feature can force the gap between almost the same bounding boxes because even if the confidence is the same, the ranking there is still an order gap. According to this order gap model is easy to learn to select the top-ranked bounding boxes to keep so that for n identical bounding boxes, the model can still distinguish the difference between them according to the rank feature, and choose the top-rank bounding box reserved.

\subsection{Greedy Matching}
Due to the limitations of one-to-one matching when dealing with nearly identical redundant bounding boxes, it is unreasonable to determine bounding box retention based on minimizing the cost of the bipartite matching matrix as in one-to-one matching. We propose a label assignment approach called greedy matching, where each bounding box is assigned a ground truth label with the minimum matching loss, and the cost matrix calculation method is consistent with stage3's one-to-one matching. We cluster the bounding boxes based on the ground truths and only retain the highest ranked bounding box in each cluster, assigning a label 1 to the retained bounding box and a label 0 to the filtered-out ones. Due to the difficulty in achieving a strict positive correlation between confidence and IoU with current detectors, this leads to situations as shown in Figure 5, where high rank does not necessarily imply good localization. Solely relying on rank for selection can affect the model's ability to choose better bounding boxes, resulting in the selection of poorly-located red bounding boxes with high scores, ultimately impairing model performance. To address this phenomenon, we introduce an $\theta$ value to adjust the label assignment. Specifically, we assign a rank of -1 to an object whose IoU with the assigned ground truth is less than y and has a rank less than $\theta$, and only retain objects with a rank greater than 1. This approach allows for more reasonable label assignments and higher performance. Additionally, by clustering bounding boxes based on their respective ground truths, our method guides the model to abandon clusters with poorly localized objects, which reduces task difficulty and accelerates the learning of filtering strategies.

\section{DEYOv2}
\subsection{Model Overview}
Our model uses YOLOv5x as the first stage, and DINO \cite{14} as the second and third stage, affording a new, progressive inference-based three-stage model. YOLOv5x is a detector of the classic YOLO series; it contains a backbone, a neck comprising FPN \cite{34} + PAN \cite{35}, and a head that outputs three-scale prediction. As a DETR-like model, DINO contains a backbone, a multi-layer transformer encoder \cite{10}, a multi-layer transformer decoder \cite{10}, and multiple prediction heads. It uses static query and dynamic initialization of the anchor bounding boxes and involves an additional CDN branch for comparative denoising training. The biggest difference between DEYOv2 and DINO is that stage3 of DEYOv2 uses high-quality proposal queries from stage2 to initialize the query. The overall DEYOv2 model is illustrated in Figure 6.

\subsection{Dense Query to Sparse Query}
Since the computational complexity of the transformer increases quadratically as the number of queries increases, the number of query-based is generally 300, 900, and taking YOLOv5 as an example, it can provide tens of thousands of queries, which is several orders of magnitude more than query-based detectors, and the cost of maintaining a single query is compared to query-based detection device is much lower. We believe that the information contained in the dense query can greatly reduce the burden of subsequent sparse query detection so that stage2 and stage3 can achieve better performance in dense detection scenarios. However, since the computational complexity of the transformer has a quadratic relationship with the sequence, this limits the number of queries in stage2 and stage3. Therefore, we use the query selector to select high-quality queries and retain the information contained in the original dense queries as much as possible in limited queries. Stage3 has fewer queries than stage1, making it possible for us to model the relationship between queries, and we can inject more information into sparse queries; even though the cost of maintaining a single sparse query is higher than that of dense query the cost can also be a good trade-off between accuracy and speed. We use a simple top-k and adapter module to complete the transition from dense query to sparse query. We believe that the adapter module can preserve the information of the original Dense Query to the greatest extent and make the dimensions of different output layer queries consistent with the dimension of the sparse query.

\begin{figure}[t]
\begin{center}
\includegraphics[width=0.9\linewidth]{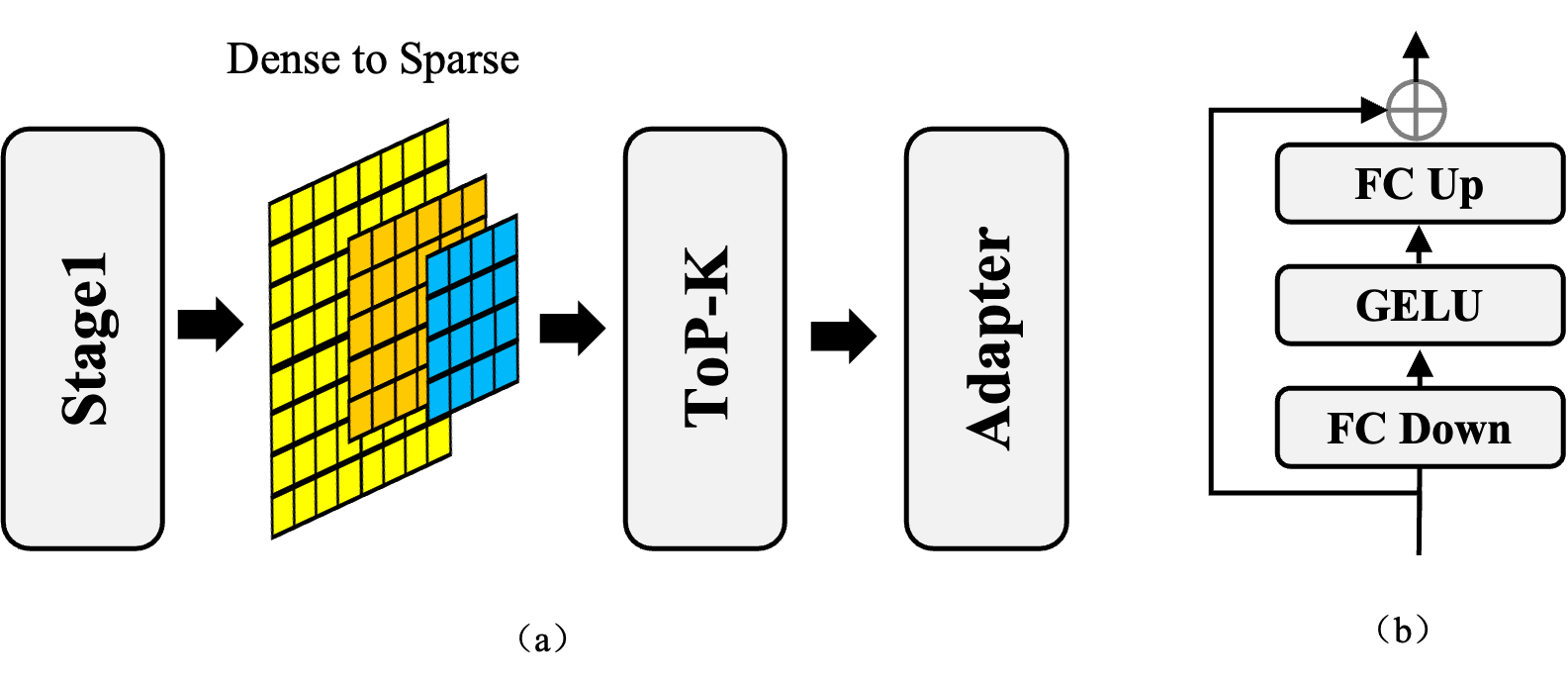}
   \caption{Details of Query Selector and adapter components. We use the index of the post-processing output of YOLOv5x to inverse solve the position of each object in the corresponding feature map.
}
\end{center}
\label{fig:7}
\end{figure}

\subsection{Query Selector}
For end-to-end optimization, we simply use top-k to complete the transition from dense query to sparse query. However, it should be noted that NMS is still a transition method to preserve stage1 information to the greatest extent. As shown in Figure 7, since there are a large number of redundant bounding boxes in stage1, directly using the top-k selection strategy will lead to a large amount of information redundancy in the selected query. NMS can solve this problem. The motivation of NMS adopted in DEYOv2 here is different from DEYO\cite{52}, just to filter out redundant information as much as possible. Due to the existence of query filter, we don't have to worry about the unreasonable setting of IoU threshold, which makes it difficult to optimize the model. This allows us to quickly adjust the IoU threshold parameters based on PRE-Matching\cite{52}, instead of evaluating the quality of the IoU threshold only after getting the final result of training like DEYO\cite{52}.

\subsection{Query Filter}
DEYOv2 uses query filter to abandon the dependence of NMS on filtering redundant bounding boxes and fully realizes end-to-end. Query filter only needs to set a confidence threshold to filter candidate bounding boxes like most end-to-end detectors to complete the filtering of redundant bounding boxes. In DEYOv2, we set the confidence threshold as 0.1. We use greedy matching to assign labels to the output of the query filter. We use focal loss to supervise the filtering of redundant bounding boxes by the query filter. Through a simple strategy: 0 means filter out, 1 means retention, and return a large number of similar redundant objects with high confidence to low confidence. Based on this strategy, the established score gap can be passed to stage3, which is convenient for stage3 to further filter out the bounding box. Since the number of filtered queries is not constant, as shown in Figure 8, we insert the filtered query into the init query of the decoder; batch training can be guaranteed.

\begin{figure}[t]
\begin{center}
\includegraphics[width=0.98\linewidth]{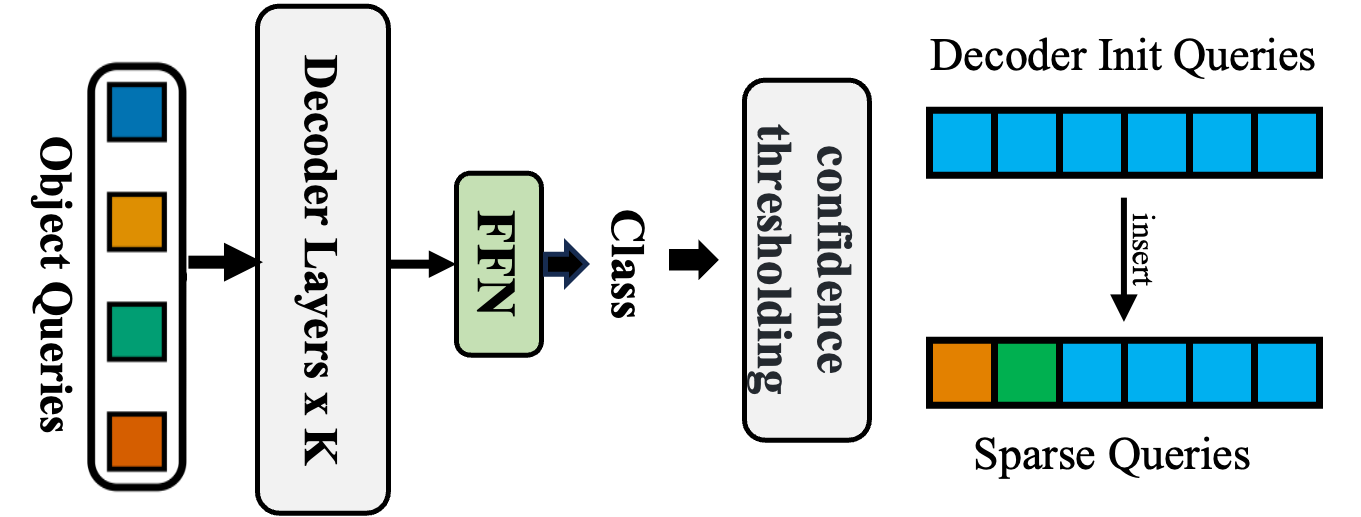}
\vspace{3mm}
   \caption{Details of Query Filter. We only establish the score gap to ensure that the score gap can be well passed to stage3.
}
\end{center}
\label{fig:8}
\end{figure}

\begin{table*}[t]
\begin{center}
\begin{tabular}{lcccccccc}
\hline
Model & Epochs & AP & AP$_5$$_0$ & AP$_7$$_5$ & AP$_S$ & AP$_M$ & AP$_L$ \\
\hline
Faster-RCNN \cite{9} &108 & 42.0 &62.4 &44.2 &20.5 &44.8 &61.1\\
DETR(DC5) \cite{1} &500 & 43.3 &63.1 &45.9 &22.5 &47.3 &61.1\\
Conditional DETR\cite{22} &108 &43.0 &64.0 &45.7 &22.7 &46.7 &61.5\\
TSP-RCNN-R \cite{39} & 96 &45.0 &64.5 &49.6 &29.7 &47.7 &58.0\\
Anchor DETR\cite{46} &50 & 42.1 &63.1 &44.9 &22.3 &46.2 &60.0\\
AdaMixer\cite{48} &36 &47.0 &66.0 &51.1 &30.1 &50.2 &61.8\\
Sparse DETR\cite{47} &50 &46.3 &66.0 &50.1 &29.0 &49.5 &60.8\\
Efficient DETR\cite{49} &36 &45.1 &63.1 &49.1 &28.3 &48.4 &59.0\\
$\mathcal{H}$-DETR\cite{50}  &12 &48.7 &66.4 &52.9 &31.2 &51.5 &63.5\\
$\mathcal{H}$-DETR\cite{50}  &36 &50.0 &68.3 &54.4 &32.9 &52.7 &65.3\\
Co-DETR\cite{51} &12 &49.5 &67.6 &54.3 &32.4 &52.7 &63.7\\
Deformable DETR \cite{7} &50 & 46.2 &65.2 &50.0 &28.8 &49.2 &61.7\\
SMCA-R \cite{38} & 50 &43.7 &63.6 &47.2 &24.2 &47.0 &60.4\\
Dynamic DETR \cite{2} &50 &47.2 &65.9 &51.1 &28.6 &49.3 &59.1\\
DAB-Deformable DETR \cite{8} &50 &46.9 &66.0 &50.8 &30.1 &50.4 &62.5\\
DN-Deformable DETR \cite{13} &50 &48.6 &67.4 &52.2 &31.0 &52.0 &63.7\\
YOLOv5x \cite{24} &-- &50.7 &-- &-- &-- &-- &--\\
DINO$_{4scale}$ \cite{52} &12 &49.0 &66.6 &53.5 &32.0 &52.3 &63.0\\
DINO$_{4scale}$  \cite{52} &24 &50.4 &68.4 &54.9 &34.0 &53.6 &64.6\\
DINO$_{4scale}$  \cite{52} &36 &50.9 &69.0 &55.3 &34.6 &54.1 &64.6\\
\hline
DEYOv2$_{4scale}$ &12 &{\bf51.1} &68.9 &55.4 &34.6 &55.2 &64.8\\
DEYOv2$_{4scale}$ &24 &{\bf51.8} &69.8 &56.3 &34.4 &56.0 &65.7\\
\hline
\end{tabular}
\end{center}
\caption{Results for DEYOv2 and other detection models with the ResNet-50 backbone(except YOLOv5x) on COCO2017 val dataset.}
\label{table:1}
\vspace{2mm}
\end{table*}

\begin{table*}[h]
\begin{center}
\begin{tabular}{lcccccccc}
\hline
Model & Epochs &NMS & AP & AP$_5$$_0$ & AP$_7$$_5$ & AP$_S$ & AP$_M$ & AP$_L$ \\
\hline
DEYO$_{4scale}$\cite{52} &12 &\checkmark &50.6 &68.7 &55.1 &33.4 &54.7 &65.3\\
DEYO$_{4scale\dagger}$\cite{52} &12 &\checkmark &51.1 &69.5 &55.4 &34.3 &55.7 &64.7\\
DEYO$_{4scale}$\cite{52} &24 &\checkmark &51.7 &70.0 &56.3 &34.6 &55.6 &65.8\\
DDQ DETR$_{4scale}$$_*$\cite{53} &12 &\checkmark &51.3 &68.6 &56.4 &33.5 &54.9 &65.9\\
DDQ DETR$_{4scale}$$_*$\cite{53} &24 &\checkmark &52.0 &69.5 &57.2 &35.2 & 54.9 &65.9\\
\hline
DEYOv2$_{4scale}$ &12 & &51.1 &68.9 &55.4 &34.6 &55.2 &64.8\\
DEYOv2$_{4scale}$ &24 & &51.8 &69.8 &56.3 &34.4 &56.0 &65.7\\
\hline
\end{tabular}
\end{center}
\caption{Compared with our most related work DDQ DETR, DEYOv2 eliminates the dependence on NMS with almost no performance loss and achieves a fully end-to-end. $\dagger$ indicates to use of the adapter, and $*$ indicates to use of an additional auxiliary head in the decoder.}
\label{table:2}
\end{table*}

\vspace{-0.9mm}
\subsection{One-to-Many with One-to-One Matching}
As analyzed earlier in this paper, the strict positive correlation between IoU and confidence is a difficult condition for current detectors. In greedy matching, we filter out all the bounding boxes with IoU \textless 0.6 with the real bounding box to make the model achieve better performance. Secondly, greedy matching clusters around the ground truth, both of which will result in the filtering of some poorly positioned bounding boxes. Query filter filters out some redundant bounding boxes at the cost of the recall rate of some poorly positioned bounding boxes, which affects the final performance to a certain extent, so we use stage3 with one-to-one matching to compensate for this part of the performance loss. At the same time, because the query filter has provided stage3 with most of the objects in the one-to-one object set, this makes the task difficulty of stage3 lower than that of the traditional DETR initialization query task. At the same time, stage3 can also further filter out bounding boxes that are not correctly filtered out by stage2. stage2 and stage3 complement each other, making DEYOv2 achieve better performance.

\section{Experiments}
\subsection{Setups}
\noindent {\bf Dataset and Backbone} We conducted the experiments on the COCO 2017 object detection dataset \cite{12}. The backbone of stage2 and stage3 is ResNet-50 \cite{15} pre-trained on ImageNet-1k, and our model is trained on the COCO 2017 training set\cite{12} without additional training data. The trained model is evaluated on the COCO 2017 validation dataset\cite{12}.
\subsection{Implementation Details}
Query filter uses a 1-layer transformer decoder, stage3 uses a 6-layer transformer encoder \cite{10}, and a 6-layer transformer decoder \cite{10} with a hidden feature dimension of 256. The model is trained using the AdamW \cite{33} optimizer with a weight decay rate of 10$^-$$^4$ and a simple learning rate adjusting strategy that decreases the learning rate to 10$^-$$^5$ at epochs 11, 20. The batch size is 16, and eight GPUs with a batch size of two each are used. Additionally, the data enhancement scheme involves random cropping and scale enhancement, and the input images are resized randomly so that the short side is between 480 and 800 pixels and the long side is up to 1333 pixels. The resized images are then padded to have a 640x640 pixels resolution and are input to the first stage of the YOLO model. Two types of images with different sizes are used in the different stages of our model for the model’s forward computation. During training, the stage1 gradients are frozen; furthermore, the gradient truncation from stage2 to stage3.

\subsection{Main Results}
As shown in Table~\ref{table:1}, we firstly compare our DEYOv2 on COCO object detection val2017 set\cite{12} with other DETR variants with ResNet-50 backbone. DEYOv2$_{4scale}$ can achieve 51.1 AP on 1 $\times$ scheduler, which gains 2.1 AP over the DINO$_{4scale}$ 1 $\times$ baselines. And with 2 $\times$ training scheduler, DEYOv2$_{4scale}$ even increased AP by 1.4 AP and 0.9 AP compared with DINO$_{4scale}$ 2 $\times$ and 3 $\times$ baselines. Table~\ref{table:2} compares DEYOv2 with DDQ DETR\cite{53}, which is the closest to our work. DEYOv2 has almost no performance loss compared to the two schedulers of DDQ DETR\cite{53} without using the auxiliary head in the decoder, and it is completely end-to-end.

\begin{table}[t]
\begin{center}
\resizebox{0.45\textwidth}{!}{
\begin{tabular}{lccccccc}
\hline
&Row &o2o & greedy &select symbol &NMS &rank feature &AP \\
\hline
&1. &\checkmark & & &  &\checkmark &0\\
&2. & &\checkmark & &  &&19.2\\
&3. & &\checkmark &\checkmark &  &\checkmark &40.7\\
&4. & &\checkmark & & &\checkmark &\bf{41.0}\\
&5. & &\checkmark & &\checkmark  &\checkmark &\bf{45.1}\\
\hline
\end{tabular}}
\end{center}
\caption{Detailed ablation of each component of DEYOv2. The results of each row are obtained by training an epoch on COCO train2017 using the ResNet5 4-scale backbone network}
\label{table:3}
\vspace{3mm}
\end{table}

\subsection{Ablation Study}
As shown in Table~\ref{table:3}, we perform the ablation study on the strong baseline DEYOv2 which exhibits 41.0 AP in the first epoch. Among them, NMS means that the query selector uses NMS instead of top-k, and select symbol uses an additional binary classification branch to assist the query filter to filter out redundant boxes. By comparing the results of 1, 2, and 3, it can be seen that greedy matching and rank feature are crucial to the filtering of dense redundant bounding boxes, and both are indispensable. By comparing 3 and 4, we found that although adding the select symbol can improve the result of stage2, the select symbol prevents the score gap from being passed to stage3, thereby degrading the performance. The results of 4 and 5 show that compared with top-k selection, NMS is still the algorithm that preserves the information of dense query to the largest extent and transitions to sparse query information. The results in Figure 9 also show that the high threshold stage1 is used to pass to stage2 The potential information is the largest. For scenarios that do not have strict requirements for end-to-end, it is also a good choice to choose high threshold (threshold = 0.9) NMS  and query filter.

\begin{table}[h]
\vspace{2mm}
\begin{center}
\resizebox{0.45\textwidth}{!}{
\begin{tabular}{lccccccc}
\hline
&Row &top-k & conf &stage2 &stage3 &AP \\
\hline
&1. & &\checkmark &\checkmark   & &49.4\\
&2. & &\checkmark &\checkmark  &\checkmark &51.8\\
&3. &\checkmark & &\checkmark &\checkmark  &41.0\\
&4. & &\checkmark &\checkmark &\checkmark   &41.0\\
\hline
\end{tabular}}
\end{center}
\caption{The ablation experiments on the effectiveness of Query Selector and stage3, 1 and 2 are obtained by using the ResNet5 4-scale backbone network to train 24 epochs on COCO train2017, and 3 and 4 are the results of the first epoch. Where k = 100, conf = 0.1.}
\label{table:4}
\end{table}

As shown in Table~\ref{table:4}, DEYOv2 also provides flexible options for the query filter to filter bounding boxes. In practical applications, top-k or confidence can be used to select bounding boxes, and better performance can be obtained by adjusting the number of reserved bounding boxes or the confidence threshold. Our experiments have obtained SOTA results, so we did not conduct further research on this parameter, which also reflects the ease of use of DEYOv2, and even a reasonable parameter can achieve amazing performance.

\subsection{Analysis}

\begin{figure}[h]
\begin{center}
\includegraphics[width=0.8\linewidth]{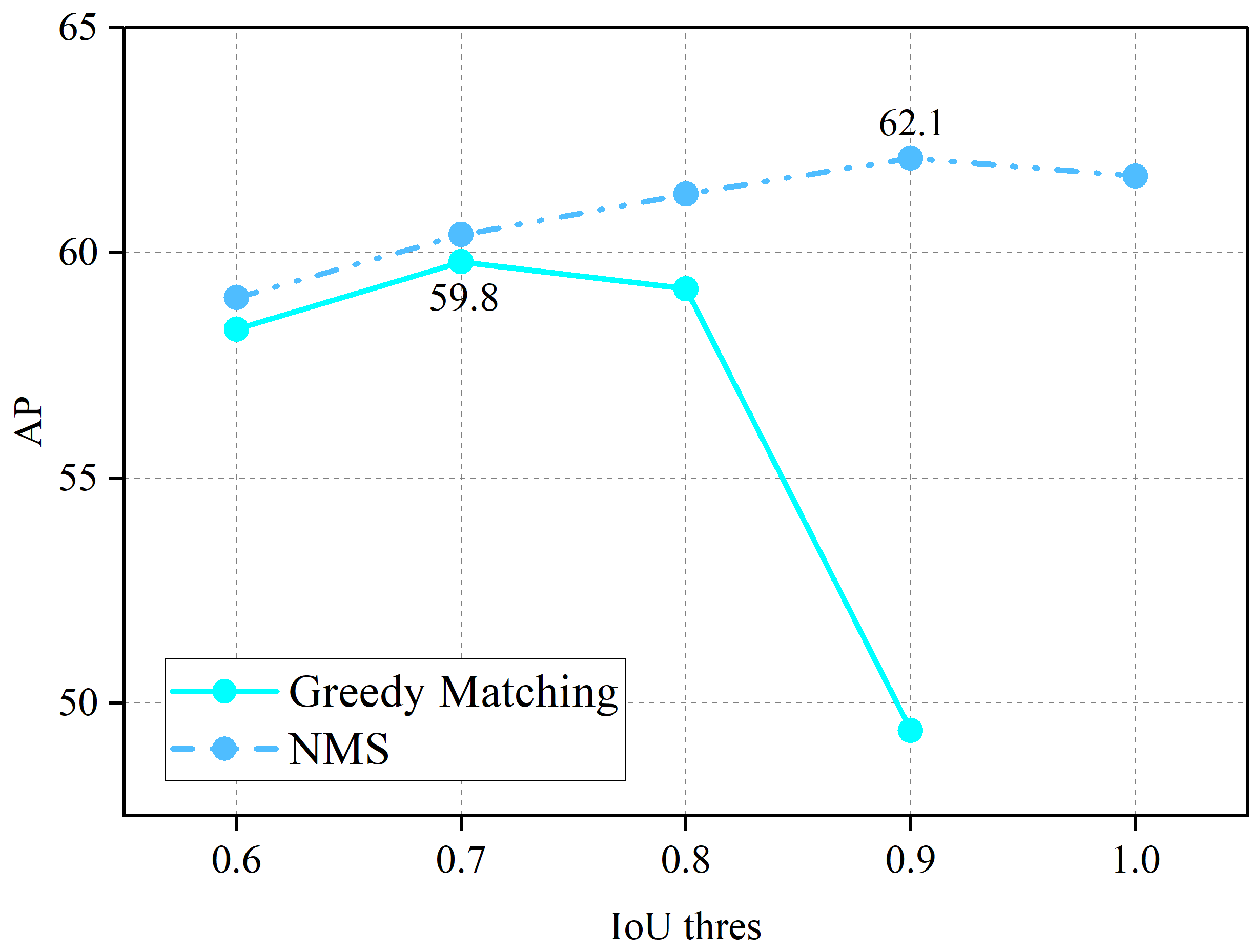}
   \caption{We compare the potential performance of NMS and greedy matching boxes using ground truth matched with the prediction set to measure label assignment ideality performance.
}
\end{center}
\label{fig:9}
\vspace{-4mm}
\end{figure}

\noindent{\bf Compare Query filter with NMS.} There is one thing in common between query filter and NMS, they both filter out redundant bounding boxes at the expense of recall. However, NMS completes the challenging step of distinguishing true positives (TP) and false positives (FP) through simple category and Bounding box information, but query filter, like one-to-matching, can combine more complex information of query, To choose a better strategy to filter out redundant bounding boxes. The recall rate sacrificed by query filter is often those poorly positioned bounding boxes, while NMS depends on the degree of object crowding in real data. The performance of the query filter will increase with the increase of detector localization performance, while NMS can only perform well in specific data. We believe that as detectors become more powerful, the performance gap between the two will become more and more obvious. One of the advantages of NMS is that using NMS is equivalent to adding a priori of a powerful filtering strategy to the detector. It can achieve good results without training, which is what query filter cannot do, but this is precisely the defect of NMS. Its inability to learn from data makes NMS one of the performance bottlenecks of the detector. With the advent of the era of large models, we believe that the performance of the query filter will get better and better as the number of model parameters and calculations increases.

\hspace*{\fill}

\begin{figure}[t]
\begin{center}
\includegraphics[width=0.8\linewidth]{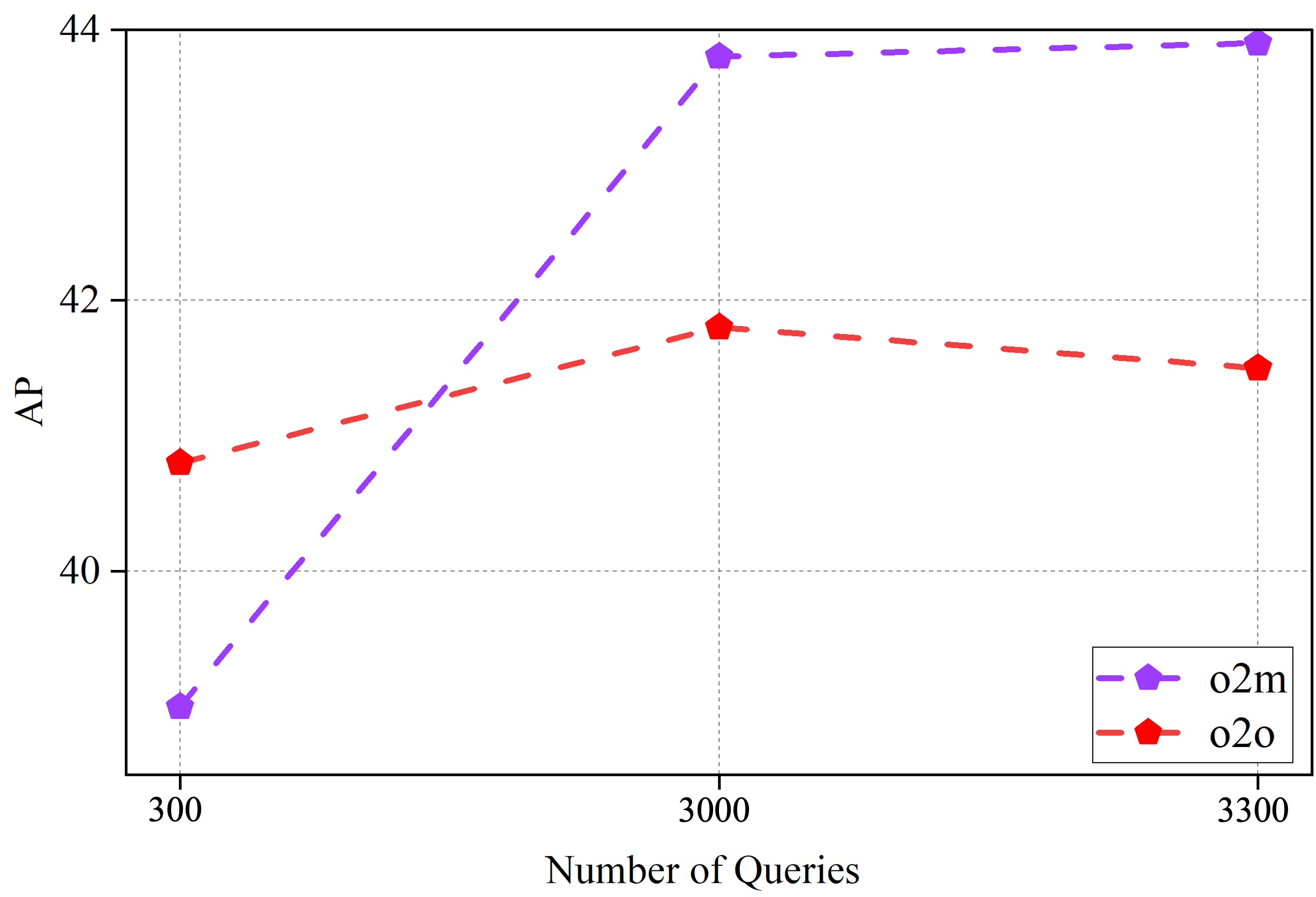}
   \caption{Results of Group DETR\cite{44} on Conditional-DETR-C5 on COCO. Increasing the number of object queries significantly affects one-to-one assignments, but not one-to-many assignments.
}
\end{center}
\label{fig:9}
\vspace{-4mm}
\end{figure}

\noindent{\bf Why is YOLOv5x?} The use of YOLOv5x is only for the convenience of verifying the effectiveness of the three stages of DEYOv2. We believe that a good three-stage paradigm detector like DEYOv2 needs to be carefully designed and weighed in each stage, but at the same time, it also has the advantage of extremely high flexibility. YOLOv5x and DINO are just a special case; our method is not designed to be compatible with all classical detectors; whether it is effective to adopt other models is beyond the scope of this paper. The implementation of DEYOv2 may not be elegant, but there is no doubt that its demonstrated performance proves the effectiveness of this paradigm. In general, DEYOv2 provides a new option for the visual community. Just like other new methods, DEYOv2 also needs to undergo the test of engineering application.

\hspace*{\fill}

\noindent{\bf Why step-by-step is better than two-stage?} Simply attributing the effectiveness of step-by-step to YOLOv5x, which can provide a large number of low-cost queries, does not touch the essence of step-by-step effectiveness. Because the encoder in DINO with two-stage configuration can also provide more than tens of thousands of queries, and since the encoder is an essential part of the entire pipeline, the advantage of low cost is not reflected on the whole. Therefore, from the perspective of the number of proposals alone, it does not reflect the advantages and necessity of step-by-step compared to two-stage. So why does using step-by-step bring such an amazing performance improvement compared to two-stage? We think that the proposal generated by one-to-one matching in two-stage is not a dense query in the true sense. Although the number of proposals can also reach tens of thousands, it is important to note that two-stage employs a one-to-one matching strategy. As shown in Figure 10, the performance deteriorates as the number of queries from the decoder increases. Excluding the performance improvement caused by the increase in the recall due to an increase in the number of queries, the performance degradation would be more severe. We can confidently assert that this phenomenon is likely to manifest in the proposal query generated by the two-stage encoder as well. However, large query numbers with one-to-many matching can avoid this problem. As shown in the Figure 10, the performance of large query numbers with one-to-many matching will not decrease as the number of queries increases, and its information density is much greater than that of large query numbers with one-to-one matching, which is the true sense of dense query. At the same time, one-to-many matching can generate more supervision signals, which can make full use of the advantages of large query numbers and accelerate convergence. Then why not directly use one-to-many matching to optimize the two-stage encoder but use step-by-step? Due to the advantages of progressive reasoning, step-by-step outperforms the two-stage approach by reducing the difficulty of the encoder's task. This allows DEYOv2 to utilize the static initialization of queries in the decoder. DEYO with adapters achieved a 48.0 AP in the first epoch, indicating that decoding proposal queries from YOLO is a relatively simple task. This enables the encoder to focus its attention on a subset of the object set, providing high-quality keys for stage3 predictions in the decoder, instead of predicting the entire object set as in the two-stage approach. Consequently, the encoding difficulty of the encoder is greatly reduced. Additionally, the step-by-step design offers more flexibility, allowing the DEYO series to benefit from advances in both classic detectors and query-based detectors (like label assignment and data augmentation, etc.).

\section{Related Work}

\noindent {\bf End-to-End Object Detector.} Relation Networks\cite{56} is the first end-to-end detection model, but the defect of this method is that it uses additional handcrafted object features. DETR\cite{1} uses the transformer to completely abandon the dependence on manual components, reduce the prior knowledge in the model, and effectively solve this problem. But DETR\cite{1} also exposed a series of problems, the most prominent of which is slow convergence. Several subsequent studies have enhanced DETR\cite{1} from various angles, effectively tackling the aforementioned issues. For example, by designing a new attention module, deformable DETR\cite{7} focuses attention on sampling points around the reference point to improve the efficiency of cross-attention. Conditional DETR\cite{22} decouples DETR queries into what and where parts to clarify the meaning of the query. Furthermore, DAB-DETR\cite{8} treats queries as 4D anchor boxes and refines them layer by layer. Based on DAB-DETR\cite{8}, DN-DETR\cite{13} believes that the slow training convergence of DETR is due to the bipartite graph matching instability in the early training stage. Therefore, the introduction of denoising group technology significantly accelerates the convergence of the DETR model. Driven by the above model, DINO\cite{52} further improved DETR\cite{1} by using Object365\cite{37}s for detection pre-training and Swin-Transformer\cite{23} as the backbone network and obtained a result of 63.3AP on the COCO val2017 dataset \cite{12}, making it transformer-based the method has become the mainstream detector for large-scale training. RT-DETR\cite{57} analyzes the reasons for the slow speed of DETR and proposes real-time DETR with real-time reasoning ability.

\hspace*{\fill}

\noindent{\bf Classic Detector with Query-Based Object Detector.} DEYO\cite{52} solves the problems of DETR itself from a novel perspective. It is the first method to combine classical detectors and DETR models. By combining the respective advantages of classical detectors and DETR-like models, DEYO achieved the most advanced results at that time. The closest work to ours is Dense Distinct Query for End-to-End Object Detection (DDQ)\cite{53}; DDQ\cite{53} combines the advantages of traditional and recent end-to-end detectors, significantly improving the performance of various detectors, including FCN, R-CNN, and DETR. However, both DEYO and DDQ\cite{53} have a common problem, and both rely heavily on NMS to filter out redundant bounding boxes from classic detectors to avoid affecting the optimization of one-to-one matching. DEYOv2 completely got rid of the dependence on NMS, and is the first fully end-to-end model that combines the respective advantages of classic detectors and query-based object Detector models.

\hspace*{\fill}

\noindent {\bf Non-Maximum Suppression.} Non-maximum Suppression is a post-processing technique commonly used in object detection to remove redundant bounding boxes. In recent years, some work has attempted to improve NMS, such as Soft-NMS\cite{30}, Softer NMS\cite{31}, and Adaptive NMS\cite{32}. However, none of these can overcome the inherent problem of NMS, which is to filter out redundant bounding boxes based only on simple category and bounding box information. The performance of the NMS algorithm largely depends on the choice of the threshold. If the threshold is set too low, too many overlapping bounding boxes will be retained, reducing the detection accuracy; if the threshold is set too high, it may lead to missed detection. At the same time, although there are some optimization methods to speed up the NMS algorithm, it may still become the bottleneck of the entire detection pipeline when a large number of detection boxes need to be processed.

\section{Conclusion}
In this paper, we propose a powerful end-to-end detector DEYOv2, that combines the advantages of classical and query-based detectors. It abandons the dependence on NMS by using the rank feature and greedy matching to achieve end-to-end optimization. The three-stage object detection paradigm of DEYOv2 has great potential in the future because of its end-to-end detection optimization effects and its excellent performance. In the future, we can apply this paradigm to lightweight detection. In the encoder, the three stages share the backbone network and use a more efficient encoder to balance performance and speed better, thereby achieving Real-time Object Detection.

Considering the limitations of this paper, before the sparse query is filtered out, we have tried \cite{50, 55} and a strategy similar to greedy matching to update sparse query using one-to-many label assignment, but the results are not as good as directly using the query from stage1. In the future, we will explore more suitable label assignments to better transition from one-to-many label assignment to one-to-many label assignment.



\vspace{0mm}
{\small
\bibliographystyle{ieee_fullname}
\bibliography{egbib}
}

\begin{figure*}[t]
\begin{center}
\includegraphics[width=0.75\linewidth]{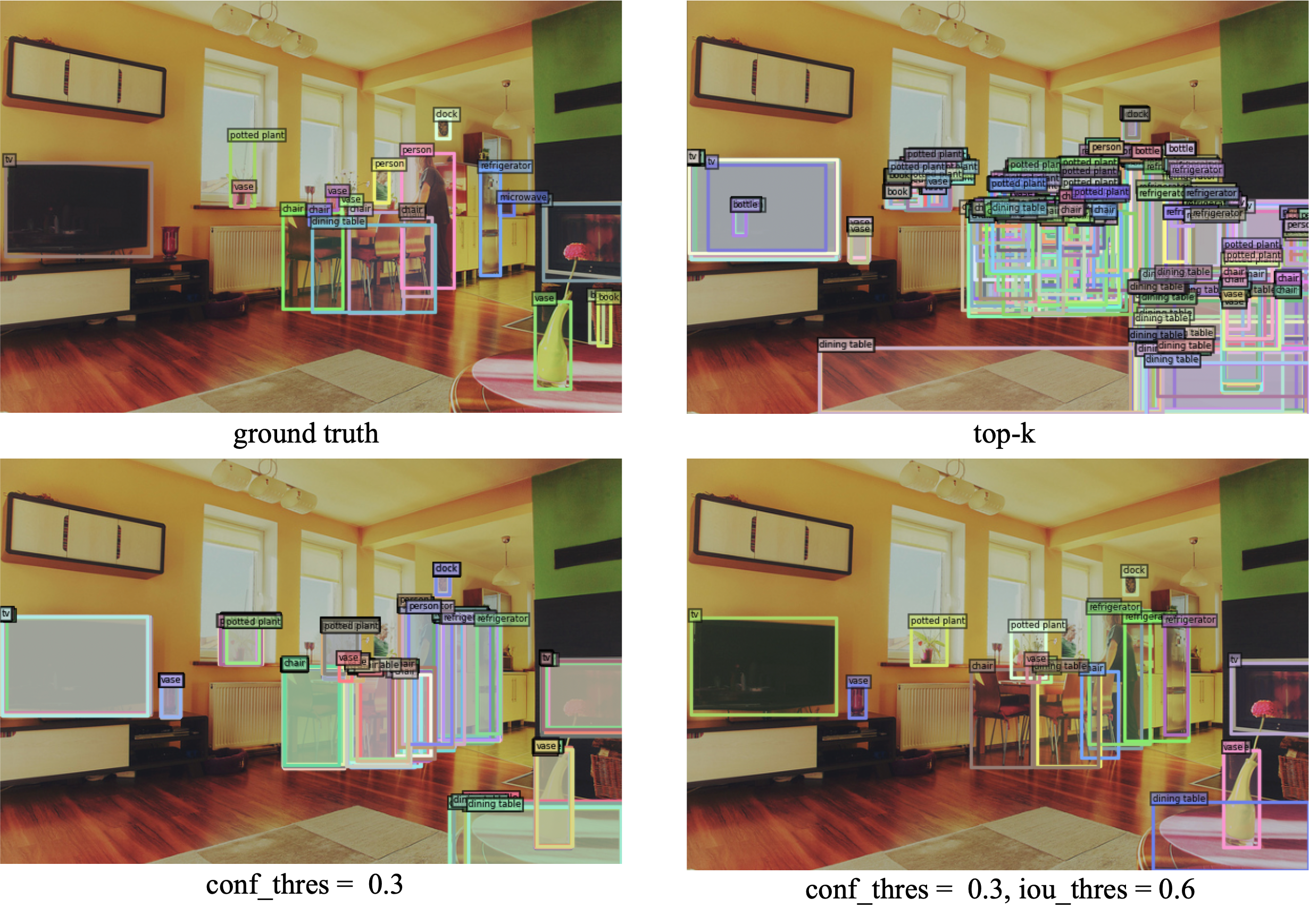}
   \caption{YOLOv5x uses different post-processing methods to visualize the results. It can be found that the original output of YOLOv5 cannot distinguish redundant bounding boxes according to the score gap.
}
\end{center}
\label{fig:10}
\end{figure*}

\begin{figure*}[h!]
\begin{center}
\includegraphics[width=0.75\linewidth]{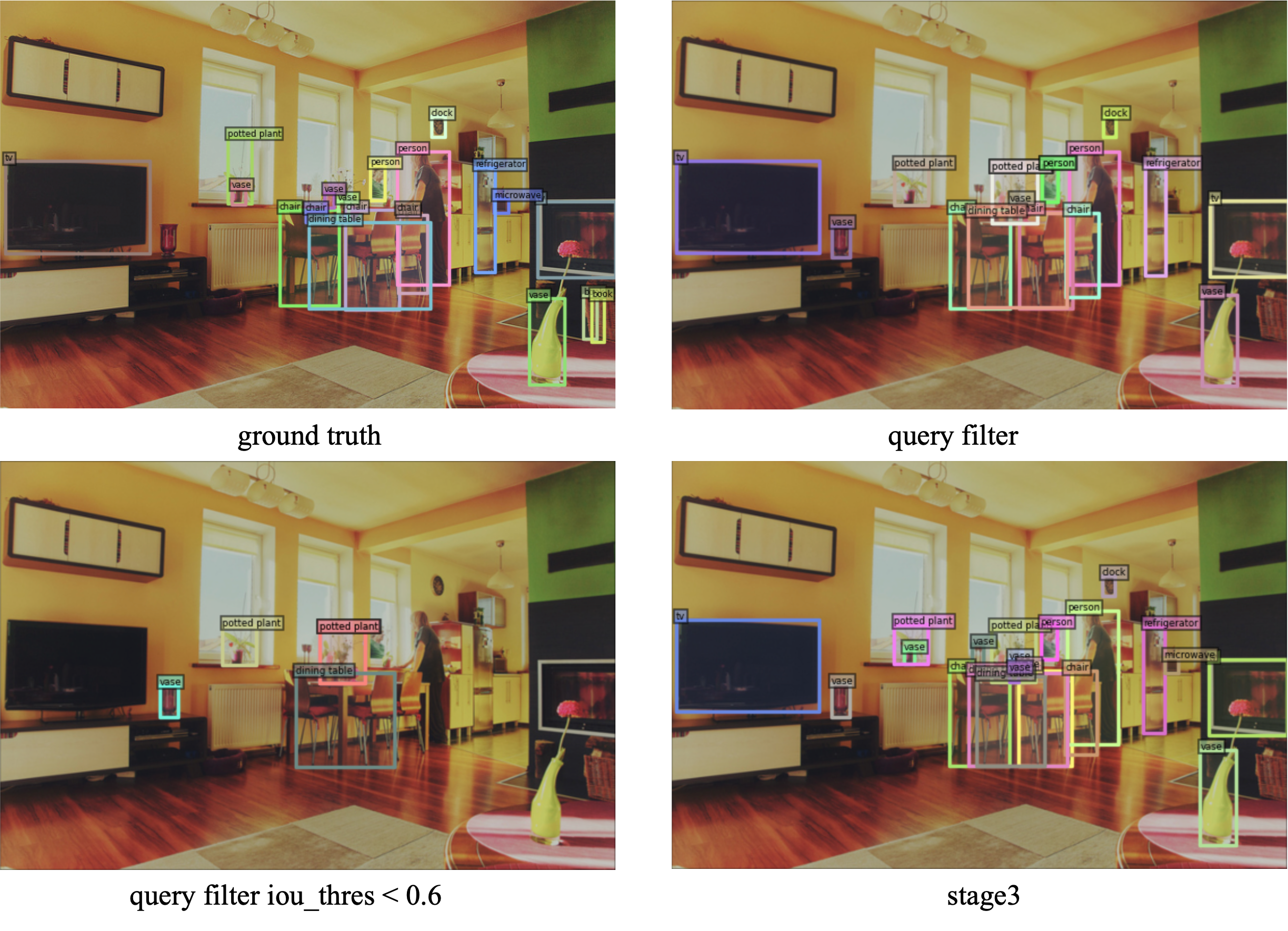}
   \caption{IoU threshold  \textless0.6 means that our output object satisfies the prediction with score \textgreater 0.3 and IoU \textless0.6 between the ground truth. It can be found that greedy matching has good generalization ability; even if the bounding box is a certain distance from the ground truth, the loss function may not give optimal supervision, but the model can still learn the filtering strategy well.
}
\end{center}
\label{fig:11}
\end{figure*}

\end{document}